\documentclass[letter]{ieice}
\usepackage[pdftex]{graphicx,xcolor}
\usepackage[fleqn]{amsmath}
\usepackage{newtxtext}
\usepackage[varg]{newtxmath}
\usepackage{url}
\usepackage{multirow}
\usepackage{lipsum} 
\usepackage{amsmath,amsfonts}
\usepackage[hidelinks]{hyperref}
\usepackage{cite}
\usepackage{algorithm}
\usepackage{array}
\usepackage{algpseudocode}
\usepackage{graphicx}
\usepackage{textcomp}
\usepackage{xcolor}
\usepackage{physics}
\usepackage{multirow}
\usepackage{subcaption}
\usepackage{mathtools}
\usepackage{tikz}
\usetikzlibrary{calc}
\usepackage{zref-savepos}
\usepackage{pict2e}
\usepackage{babel}

\usepackage{pifont}
\usepackage[outline]{contour}

\DeclarePairedDelimiter\ceil{\lceil}{\rceil}
\DeclarePairedDelimiter\floor{\lfloor}{\rfloor}

\newcolumntype{M}[1]{>{\centering\arraybackslash}m{#1}}

\newcommand\blfootnote[1]{%
  \begingroup
  \renewcommand\thefootnote{}\footnote{#1}%
  \addtocounter{footnote}{-1}%
  \endgroup
}

\field{D}
\title{Hybrid Temporal-8-Bit Spike Coding for Spiking Neural Network Surrogate Training}
\authorlist{%
 \authorentry{Nhan T. Luu}{m}{a1}\MembershipNumber{}
 \authorentry{Duong T. Luu}{n}{a2}\MembershipNumber{}
 \authorentry{Nam Pham Ngoc}{n}{a3}\MembershipNumber{}
 \authorentry{Thang Truong Cong}{m}{a4}\MembershipNumber{}
}
\affiliate[a1]{This author is with College of Information and Communication Technology, Can Tho University, Vietnam and College of Engineering and Computer Science, VinUniversity, Hanoi, Vietnam (luutn@ctu.edu.vn).}
\affiliate[a2]{This author is with Center of Digital Transformation and Communication, Can Tho University, Vietnam (luutd@ctu.edu.vn).}
\affiliate[a3]{This author is with College of Engineering and Computer Science, VinUniversity, Hanoi, Vietnam (nam.pn@vinuni.edu.vn).}
\affiliate[a4]{This author is with Department of Computer Science and Engineering, The University of Aizu, Japan (thang@u-aizu.ac.jp).}

\received{2000}{11}{11}
\revised{2000}{11}{11}

\begin{document}
\maketitle
\begin{summary}
Spiking neural networks (SNNs) have emerged as a promising direction in both computational neuroscience and artificial intelligence, offering advantages such as strong biological plausibility and low energy consumption on neuromorphic hardware. Despite these benefits, SNNs still face challenges in achieving state-of-the-art performance on vision tasks. Recent work has shown that hybrid rate–temporal coding strategies (particularly those incorporating bit-plane representations of images into traditional rate coding schemes) can significantly improve performance when trained with surrogate backpropagation. Motivated by these findings, this study proposes a hybrid temporal–bit spike coding method that integrates bit-plane decompositions with temporal coding principles. Through extensive experiments across multiple computer vision benchmarks, we demonstrate that blending bit-plane information with temporal coding yields competitive, and in some cases improved, performance compared to established spike-coding techniques. To the best of our knowledge, this is the first work to introduce a hybrid temporal–bit coding scheme specifically designed for surrogate gradient training of SNNs.
\end{summary}
\begin{keywords}
Spiking neural network, temporal spike coding, image classification, bit planes
\end{keywords}
\blfootnote{Corresponding to: Nhan T. Luu.}
\section{Introduction}
The field of artificial intelligence (AI) has experienced substantial progress in recent decades, much of it driven by advances in deep learning algorithms \cite{wani2020advances}. However, conventional artificial neural networks (ANNs) \cite{goodfellow2016deep} still struggle to capture the intricate, dynamic behaviors characteristic of biological neural systems. To bridge this gap, spiking neural networks (SNNs) have emerged as a promising alternative, incorporating temporal neural dynamics to more faithfully emulate biological computation.

In contrast to traditional ANNs, SNNs communicate through discrete spike events, mirroring the asynchronous and event-driven signaling observed in biological neurons \cite{auge2021survey}. This spike-based computation not only increases biological plausibility but also enables greater computational efficiency and reduced energy consumption, particularly on neuromorphic hardware platforms \cite{blouw2019benchmarking}. The inherent sparsity and temporal coding mechanisms in SNNs form a strong foundation for the development of low-power systems capable of real-time operation \cite{rajendran2019low}.

Despite these advantages, SNNs introduce unique challenges, especially in training and optimization \cite{deng2020rethinking}. Standard backpropagation techniques developed for ANNs cannot be directly applied due to the discrete, non-differentiable nature of spikes \cite{nunes2022spiking}. Consequently, several alternative training strategies such as spike-timing-dependent plasticity (STDP) \cite{srinivasan2017spike}, ANN-to-SNN (ANN2SNN) conversion pipelines \cite{rueckauer2017conversion}, and surrogate gradient methods \cite{fang2021deep} have been proposed to make effective learning in SNNs feasible.

Recent studies have further explored enriching surrogate-based SNN training by incorporating bit-plane representations \cite{luu2024}, demonstrating improvements in tasks such as image recognition and processing. Motivated by these developments, we propose a new hybrid temporal-bit spike coding scheme that integrates temporal coding principles with multi-bit plane decomposition. Our extensive experiments show that combining temporal coding with bit-plane information yields competitive, and in several cases superior, performance compared to existing spike coding approaches. Our contributions are summarized as follows:
\begin{itemize}
    \item Building upon prior work \cite{luu2024}, we introduce a hybrid temporal-8-bit spike-coding method tailored for surrogate-gradient training of SNNs.
    \item We show that integrating bit-plane information with temporal coding achieves competitive or improved performance relative to established spike coding strategies.
    \item We demonstrate that the proposed method generalizes effectively across well-known SNN architectures and remains compatible with modern optimization algorithms widely used in SNN training.
\end{itemize}

\section{Methodology}

The decomposition of a multi-channel image into its corresponding bit planes is performed by isolating each bit in the binary representation of the pixel intensities for every color channel. For an $n_{\text{bit}}$-bit image, let $I_{C}(x,y)$ denote the intensity at spatial location $(x,y)$ in channel $C$. Its binary expansion can be written as
\begin{equation}
    I_{C}(x, y) = \sum_{k=0}^{n_{\text{bit}}} B_{k,C}(x, y)\, 2^{k},
\end{equation}
where $B_{k,C}(x,y) \in \{0,1\}$ denotes the $k$-th bit of the binary representation of $I_{C}(x,y)$, with $k=0$ corresponding to the least significant bit (LSB) and $k=n_{\text{bit}}$ corresponding to the most significant bit (MSB). The $k$-th bit plane can be extracted using
\begin{equation}
    B_{k,C}(x, y) = \left\lfloor \frac{I_{C}(x,y)}{2^{k}} \right\rfloor \bmod 2.
\end{equation}

\begin{algorithm}[t!]
\caption{Bit plane extraction algorithm}
\label{bitplane_alg} 
\begin{algorithmic}
\Require $X_{max}$ is defined, $X_{max} > 0$ and $X \in \mathbb{Z}$
\Function{BitplaneEncode}{$X, X_{max}$, $mode="MSB"$}
    \State $ n_{bit} \gets \ceil*{\log_2(X_{max})}$ \Comment{Get number of bit required to encode image from highest possible value $X$}
    \State $ L \gets [\text{ }]$ \Comment{Empty list initialization}
    
    \For{each $X_{shape}.C$}
        \For{$i \gets 0, i < n_{bit}, i++$}
            \State $L.insert(X \mod{2})$ \Comment{Element-wise modulo of tensor, then store the bit plane to $L$}
            \State $X \gets \floor*{X/2}$ \Comment{Element-wise division of input image tensor and floor the result}
        \EndFor
    \EndFor
    \If{$mode == "MSB"$}
        \State $L.reverse()$ \Comment{Reverse original LSB order to MSB format}
    \EndIf
    \State $B \gets concat(L)$ \Comment{Concatenate all stored bit planes in $L$ for model inference}
    \State \Return $B$
\EndFunction
\end{algorithmic}
\end{algorithm}

To encode an input image batch $X$ into its $n_{\text{bit}}$ bit-plane representation during training, we follow the procedure outlined in Algorithm~\ref{bitplane_alg}. For an unnormalized integer-valued tensor batch $X$, each bit plane is obtained through iterative element-wise division by 2. The remainder at each step corresponds to one bit plane, while the quotient is retained for extraction of subsequent planes. Repeating this process $n_{bit}$ times yields $n_{bit}$ binary tensor batches, each matching the original spatial dimensions $X_{shape}$. These bit planes are concatenated to form the full bit-plane representation in LSB order; optionally, the ordering may be reversed to produce an MSB format.

After obtaining all bit planes $B$, we encode the corresponding image intensities using the time-to-first-spike (TTFS) scheme. In TTFS encoding, the spike time is inversely related to the input magnitude, decreasing linearly as the intensity increases. Formally, for a normalized input value $x \in X_{normalized}$ where $X_{normalized} = X/X_{max}$ and $x \in [0, 1]$, the firing time is given by
\begin{equation}
    t_f(x) = (T - 1)(1 - x),
    \label{eq:ttfs}
\end{equation}
where $T$ denotes the maximum allowable firing time. Thus, larger input values yield earlier spikes (smaller $t_f$), while smaller values produce later spikes. To construct the proposed hybrid temporal-bit spike representation, we concatenate the TTFS-encoded spike times with the extracted bit-plane tensors. Let $X_{TTFS}$ denote the TTFS spike-timing tensor derived from the input image, and let $B = [B_0, B_1, \ldots, B_{n_{bit}-1}]$ represent the set of extracted bit planes. The final spike-chain representation is defined as
\begin{equation}
    S = \operatorname{concat}\!\left( X_{TTFS}, B_0, B_1, \cdots, B_{n_{bit-1}}\right),
    \label{eq:spike_concat}
\end{equation}
where $\operatorname{concat}(\cdot)$ denotes timestep-wise concatenation along the temporal dimension $T$.

\section{Experiments and results}

\subsection{Experimental settings}

To optimize the SNN components of our model, we employed hard-reset IF neurons with arctangent surrogate gradient function with backpropagate through time (BPTT)~\cite{fang2021deep} for all benchmarked SNN variants as:
\begin{equation}
    \begin{aligned}
        & \begin{aligned}
            u_i^{(l)}(t) = & \bigl(1 - s_i^{(l)}(t-1)\bigr)\,u_i^{(l)}(t-1)\\
            & + W^{(l)} s^{(l-1)}(t) + b_i^{(l)},\\
        \end{aligned}\\
        &s_i^{(l)}(t) = \Theta\left(u_i^{(l)}(t), V_{\text{th}}\right)\\
        & \Theta(x, V_{th}) =
        \begin{cases}
        1 & \text{if } x - V_{th} \geq 0 \\
        0 & \text{otherwise}
        \end{cases} \\
        & \nabla_x \Theta  = \frac{\alpha}{2\left( 1 + \left( \frac{\pi}{2} \alpha x \right)^2 \right)} \quad \text{where $\alpha=2$}\\
        & \frac{\partial \mathcal{L}}{\partial W^{(l)}} = \sum_{t=1}^{T} \frac{\partial \mathcal{L}}{\partial s^{(l)}(t)} \cdot \frac{\partial s^{(l)}(t)}{\partial u^{(l)}(t)} \cdot \frac{\partial u^{(l)}(t)}{\partial W^{(l)}}\\ 
    \end{aligned}
    \label{heviside}
\end{equation}
where $u_i^{(l)}(t)$ is the membrane potential of neuron $i$ in layer $l$ at time $t \in T$, $s_i^{(l)}(t)$ is the spike output of neuron $i$, similarly with prior layer signal $s_i^{(l-1)}(t)$, $V_{\text{th}}$ is the threshold voltage, synaptic weight $W^{(l)}$ and bias $b_i^{(l)}$, $\Theta$ is the Heaviside function and it's surrogate gradient $\nabla_x \Theta$, loss function $\mathcal{L}$ and it's gradient with respect to weight $\frac{\partial \mathcal{L}}{\partial W^{(l)}}$.

All experiments were conducted using the PyTorch framework~\cite{paszke2019pytorch}, along with the SpikingJelly library~\cite{doi:10.1126/sciadv.adi1480}. Inspired by traditional deep learning approach in model fine-tunning \cite{luu2023blind} and prior work in surrogate SNN optimization \cite{fang2021deep, luu2024}, optimization for all SNN parameters $\theta^\ast$ was carried out under cross-entropy loss $\mathcal{L}_{CE}$ as:
\begin{equation}
    \begin{aligned}
        \theta^\ast &= \arg\min_\theta \sum_{n=1}^N \sum_{k=1}^K -y_{nk} \log f(x_n ; \theta)_k \\
        &= \arg\min_\theta \mathcal{L}_{CE}(y, \hat{y}),
    \end{aligned}
    \label{eq:cel}
\end{equation}
along with the Adam optimizer~\cite{kingma2014adam}, with a fixed learning rate of \( lr = 10^{-3} \), decay rate of \( \lambda = 10^{-3} \) and momentum parameters \( \beta = (0.9, 0.999) \), consistently applied across all experiments. To ensure reproducibility and consistency with prior researches, we followed the original dataset partitioning wherever available. For datasets lacking predefined splits, we adopted an 80/20 training-validation split. All experiments were conducted using fixed random seeds for reproducibility, and training was carried out over 100 epochs on an NVIDIA RTX 3090 GPU with 24 GB of VRAM.

\subsection{Benchmarking with grayscale datasets}

\begin{table}[t!]
    \centering
    \caption{Comparison of SEW-RN18 cross-dataset validation accuracy among coding methods. Best performances on each dataset are denoted in bold and second best are underlined. Performance gain when compare with its equivalent baseline is denoted with green on right side of proposed method results.}
    \resizebox{\linewidth}{!}{
    \begin{tabular}{M{1cm}*{3}{M{0.8cm}}*{2}{M{1.6cm}}}
    \hline
    \multirow{3}{*}{Dataset} & \multicolumn{5}{c}{Top 1 accuracy per coding method (\%)} \\
    \cline{2-6}
     & Rate & TTFS & 8-bit & Hybrid rate-8-bit & \bf Proposed method\\
    \hline
    MNIST & 98.62 & 98.74 & 97.87 & \textbf{98.93} (\textcolor{green}{+0.31})& \underline{98.83} (\textcolor{green}{+0.09})\\
    \hline
    KMNIST & 94.13 & 93.26 & 92.33 & \underline{95.58} (\textcolor{green}{+1.45})& \textbf{95.65} (\textcolor{green}{+2.39})\\
    \hline
    FMNIST & 88.83 & 89.21 & 86.66 & \textbf{90.19} (\textcolor{green}{+1.36})& \underline{90.11} (\textcolor{green}{+1.00})\\
    \hline
    \textbf{Average} & 93.86 & 93.74 & 92.29 & \textbf{94.90} (\textcolor{green}{+1.04})& \underline{94.86} (\textcolor{green}{+1.12})\\
    \hline
    \end{tabular}}
    \label{tab:grayscale_acc_compare}
\end{table}

First, we proceed to perform a cross-dataset evaluation of our proposed method using SEW-ResNet18 (SEW-RN18)~\cite{fang2021deep} against other well spike coding strategies, including: rate coding, TTFS, bit plane coding and hybrid rate-8-bit~\cite{luu2024} across 3 benchmarked datasets (MNIST~\cite{deng2012mnist}, KMNIST~\cite{clanuwat2018deep} and Fashion-MNIST (FMNIST)~\cite{xiao2017fashion}). The results are reported in Table~\ref{tab:grayscale_acc_compare}. 

Although the accuracy gain on MNIST is relatively small (around 0.09\%), the proposed method still maintains competitive performance, either being the best or second best on all 3 dataset. Notably, our method is able to achieve the highest performance (95.65\%) on MNIST and the highest total accuracy gain (2.39\%) with respect to it corresponding baseline. On average, the proposed method improves accuracy by 1.09\% over TTFS baseline (0.05\% higher than prior hybrid rate-8-bit approach), demonstrating its consistent robustness across datasets.

\subsection{Benchmarking with different architecture and optimization algorithm}

\begin{table}[t!]
    \centering
    \caption{Comparison of CIFAR10 validation accuracy using TTFS and proposed method on different SOTA models along with different optimization algorithm. Best performances on each dataset are denoted in bold and second best are underlined. Performance gain when compare with baseline using the same optimization algorithm is denoted with green on right side of proposed method results.}
    \resizebox{\linewidth}{!}{
    \begin{tabular}{M{2.3cm}*{2}{M{0.9cm}}*{2}{M{1.7cm}}}
    \hline
    \multirow{3}{*}{Model variants} & \multicolumn{4}{c}{Top 1 accuracy per setting (\%)} \\
    \cline{2-5}
     & TTFS + Adam & TTFS + SGD & \textbf{Proposed} + Adam & \textbf{Proposed} + SGD\\
    \hline
    SEW-RN18-ADD & 72.24 & 72.36 & \textbf{74.29} (\textcolor{green}{+2.05}) & \underline{73.35} (\textcolor{green}{+0.99})\\
    \hline
    SEW-RN18-AND & 64.58 & 70.25 & \underline{72.95} (\textcolor{green}{+8.37})& \textbf{74.15} (\textcolor{green}{+3.90})\\
    \hline
    SEW-RN18-IAND & 69.28 & \underline{73.93} & 73.40 (\textcolor{green}{+4.12}) & \textbf{74.08} (\textcolor{green}{+0.15})\\
    \hline
    Spiking-RN18 & 48.15  & 58.67 & \underline{63.79} (\textcolor{green}{+15.1})& \textbf{64.93} (\textcolor{green}{+6.26})\\
    \hline
    Spiking-RN34 & 21.28 & 15.52 & \textbf{34.29} (\textcolor{green}{+13.0})& \underline{33.07} (\textcolor{green}{+17.6})\\
    \hline
    Spiking-RN50 & 19.94 & 16.03 & \textbf{29.78} (\textcolor{green}{+9.84})& \underline{22.22} (\textcolor{green}{+6.19})\\
    \hline
    \textbf{Average} & 49.25 & 51.12 & \textbf{58.08} (\textcolor{green}{+8.83})& \underline{56.97} (\textcolor{green}{+5.85})\\
    \hline
    \end{tabular}}
    \label{tab:opt_var_compare}
\end{table}

Aside from benchmarking against multiple vision dataset, we also evaluated our spike coding scheme against state-of-the-art (SOTA) spiking architecture variants under different optimization algorithms on CIFAR10. Specifically, we benchmarked against multiple Spiking-ResNet (Spiking-RN) variants with LIF surrogate neurons~\cite{hu2021spiking}, which have demonstrated competitive performance in prior work, along with AND and IAND variants of SEW-RN18 under adaptive optimization algorithm Adam \cite{kingma2014adam} and traditional stochastic gradient descent (SGD).

In Table \ref{tab:opt_var_compare}, our results reveal that in almost all of the tested configurations, proposed method would consistently achieve a higher accuracy than baseline, with both best and second best result are trained using proposed method (except for SGD optimized SEW-RN18-ADD, Spiking-RN34 and Spiking-RN34). Performance gain range from 0.15\% in the case of SGD optimized SEW-RN18-IAND to 17.6\% in the case of SGD optimized Spiking-RN34. Both selected optimizer work well with our method while evenly share best performing results, although our method work more consistently with Adam, where lowest accuracy gain is 1.90\% (as indicated in the averaged performance gain in Table \ref{tab:opt_var_compare}). Also, Spiking-RN tends to underperform on direct spike-based optimization, since its variants are meant to be used in ANN2SNN conversion~\cite{hu2021spiking}, along with problems related to vanishing gradients as mentioned in prior work \cite{fang2021deep}.

\subsection{Benchmarking with intermediate and large scale image datasets}

\begin{table}[t!]
    \centering
    \caption{Average validation accuracy of SEW-RN18 on various computer vision dataset of our proposed method. Best performances on each dataset are denoted in bold and second best are underlined. Performance gain when compare with its equivalent baseline is denoted with green on right side of proposed method results.}
    \resizebox{\linewidth}{!}{
    \begin{tabular}{M{1cm}*{3}{M{0.8cm}}*{2}{M{1.7cm}}} 
    \hline
     \multirow{3}{*}{Dataset} & \multicolumn{5}{c}{Top 1 accuracy per coding method (\%)} \\
    \cline{2-6}
     & Rate & TTFS & 8-bit & Hybrid rate-8-bit & \bf Proposed method\\
    \hline
    CIFAR10 & 70.69 & 72.24 & 37.28 & \underline{73.49} (\textcolor{green}{+2.80})& \textbf{74.29} (\textcolor{green}{+2.05})\\
    \hline
    CIFAR100 & 38.57 & 38.95 & 11.28 & \textbf{42.15} (\textcolor{green}{+3.58})& \underline{41.65} (\textcolor{green}{+2.70})\\
    \hline
    Caltech101  & 61.67 & 61.21 & 43.28 & \underline{64.55} (\textcolor{green}{+2.88})& \textbf{64.72} (\textcolor{green}{+3.51})\\
    \hline
    Caltech256 & 30.04 & 33.80 & 21.75 & \underline{41.51} (\textcolor{green}{+11.5}) & \textbf{42.29} (\textcolor{green}{+8.49})\\
    \hline
    EuroSAT & 84.39 & 87.13 & 79.87 & \textbf{88.48} (\textcolor{green}{+4.09})& \underline{87.75} (\textcolor{green}{+0.62})\\
    \hline
    Imagenette & 72.99 & 76.99 & 48.20 & \underline{78.68} (\textcolor{green}{+5.69})& \textbf{79.15} (\textcolor{green}{+2.16})\\
    \hline
    Food101 & 19.62 & 17.95 & 6.29 & \underline{36.95} (\textcolor{green}{+17.3})& \textbf{37.59} (\textcolor{green}{+19.6})\\
    \hline
    \textbf{Average}  & 53.99 & 55.47 & 35.42 & \underline{60.83} (\textcolor{green}{+6.84})& \textbf{61.06} (\textcolor{green}{+5.59})\\
    \hline
    \end{tabular}}
    \label{tab:rgb_acc_compare}
\end{table}

For our experiment, we conducted tests on SEW-RN18 over various color-image classification datasets, including CIFAR10~\cite{alex2009learning}, CIFAR100~\cite{alex2009learning}, EuroSAT~\cite{helber2019eurosat}, Caltech101~\cite{li_andreeto_ranzato_perona_2022}, Caltech256~\cite{griffin_holub_perona_2022}, Imagenette~\cite{imagenette} and Food101~\cite{food101}. The accuracy results of different cases are shown in Table \ref{tab:rgb_acc_compare}. We can see that, the accuracy of both the proposed method is always higher than its corresponding baseline, where proposed method with TTFS perform better on both total performance and relative accuracy gain with respect to baseline method. The gain ranges from the lowest of about 0.62\% (on EuroSAT dataset) to about 19.6\% (on Food101 dataset). Results also showed that our proposed method also consistently perform better than prior hybrid rate-8-bit approach by a factor of 0.23\% on average. This suggests that the benefit of the proposed method when applying to color images is more significant than when applying to grayscale images.

\section{Conclusion}

In this paper, we have introduced a hybrid temporal-8-bit coding method for SNN that leverages the combination of temporal coding signal and bit planes of input image data. Through extensive experimental validation, we have demonstrated the effectiveness of our coding strategy with multiple well known coding scheme across different grayscale and color datasets. This study also showed that our proposed method also generalize well across well-studied SNN architectures and optimization algorithm. We expect that the findings in this paper open new avenues for the development of more efficient and effective SNN models.

\section*{Data availability}
Source code for all hybrid coding method is publicly available on GitHub (\url{https://github.com/luutn2002/bit-plane-snn}).
\bibliographystyle{ieicetr}
\bibliography{refs}

\end{document}